\documentclass{article}
\usepackage{spconf,amsmath,amssymb}
\usepackage{microtype}
\usepackage{graphicx,epsfig}
\usepackage[linesnumbered,boxed]{algorithm2e}
\usepackage[caption=false,font=footnotesize]{subfig}

\def\i{\boldsymbol{i}}
\def\j{\boldsymbol{j}}
\def\Z{\mathbb{Z}}

\newtheorem{theorem}{Theorem}[section]
\newtheorem{proposition}[theorem]{Proposition}

\title{Fast and Accurate Bilateral Filtering using Gauss-Polynomial Decomposition}
\name{Kunal N. Chaudhury\thanks{K.~N.~Chaudhury was partially supported by a Startup Grant provided by the Indian Institute of Science. Correspondence: kunal@ee.iisc.ernet.in.}}
\address{Department of Electrical Engineering, Indian Institute of Science, Bangalore, India}

\begin{document}
\ninept

\maketitle

\begin{abstract}
The bilateral filter is a versatile non-linear filter that has found diverse applications in image processing, computer vision, computer graphics, and computational photography.
A common form of the filter is the Gaussian bilateral filter in which both the spatial and range kernels are Gaussian.
A direct implementation of this filter requires $O(\sigma^2)$ operations per pixel, where $\sigma$ is the standard deviation of the spatial Gaussian. In this paper, we propose an accurate approximation algorithm that can cut down the computational complexity to $O(1)$ per pixel for any arbitrary $\sigma$ (constant-time implementation). This is based on the observation that the range kernel operates via the translations of a fixed Gaussian over the range space, and that these translated Gaussians can be accurately approximated using the so-called Gauss-polynomials. The overall algorithm emerging from this approximation involves a series of spatial Gaussian filtering, which can be efficiently implemented (in parallel) using separability and recursion. We present some preliminary results to demonstrate that the proposed algorithm compares favorably with some of the existing fast algorithms in terms of speed and accuracy. 
\end{abstract}


\section{Introduction}
\label{sec:intro}

The bilateral filter of Tomasi and Maduchi \cite{Tomasi1998} is a particular instance of an edge-preserving smoothing filter. The origins of the filter can be traced back to the work of Lee \cite{Lee1983} and Yaroslavsky \cite{Yaroslavsky1985}. The SUSAN framework of Smith and Brady \cite{Smith1997} is also based  on a similar idea. The relation between the bilateral and other closely related filters is surveyed in \cite{Elad2002}. The bilateral filter has turned out to be a versatile tool that has found widespread applications in image processing, computer graphics, computer vision, and computational photography. A detailed survey of some of these applications can be found in \cite{Book2009}. More recently, the bilateral filter has received renewed attention in the context of image denoising \cite{Knaus2014,Morel2014}. 
The original bilateral filter \cite{Tomasi1998} has a straightforward extension to signals of arbitrary dimension and, in particular, to video and volume data \cite{Book2009}. Thus, while we will limit our discussion to images in this paper, the ideas that we present next can also be extended to higher-dimensional signals.

Consider a discrete image $\{ f(\i) : \i \in I\}$ where $I$ is some finite rectangular domain of $\Z^2$. The Gaussian bilateral filtering of this image is given by 
\begin{equation}
\label{BF}
 f_{\mathrm{BF}}(\i)=  \frac{\sum_{\j \in \Omega} g_{\sigma_s}(\j) \  g_{\sigma_r}(f(\i-\j)-f(\i)) \ f(\i-\j)}{\sum_{\j \in \Omega} g_{\sigma_s}(\j)  \  g_{\sigma_r}(f(\i-\j)-f(\i)) },
\end{equation}
where both the \textit{spatial}  and \textit{range} kernels are Gaussian, 
\begin{equation}
\label{kernel}
g_{\sigma_s}(\i) = \exp\left(- \frac{\lVert \i \rVert^2}{2\sigma_s^2}\right)  \quad \text{and}  \quad  g_{\sigma_r}(t) = \exp\left(- \frac{t^2}{2\sigma_r^2}\right).
\end{equation}
In practice, the domain of the spatial kernel $\Omega$ is restricted to some neighbourhood of the origin. Typically, $\Omega$ is a square neighbourhood, $\Omega=[-W,W] \times [-W,W]$ where $W=3\sigma_s$ \cite{Tomasi1998}. 
We refer the reader to \cite{Tomasi1998,Book2009} for a detailed exposition on the working of the filter.


\subsection{Fast Bilateral Filter}

It is clear that a direct implementation of \eqref{BF} requires $O(\sigma_s^2)$ operations per pixel. In general, the directly computed bilateral filter  is slow for practical settings of $\sigma_s$ \cite{Book2009}. To address this issue, researchers have come up with several fast algorithms \cite{Durand2002,Paris2006,Weiss2006,Porikli2008,Yang2009,Adams2009,Adams2010,Chaudhury2011} that are based on some form of approximation and yield various levels of speed and accuracy. 
We refer the interested reader to  \cite{Book2009,Adams2010} for a survey of  algorithms for fast bilateral filtering.
The ultimate goal in this regard is to reduce the complexity to $O(1)$ per pixel, that is, the run time of the implementation should not depend on $\sigma_s$. This is commonly referred to as a constant-time implementation. 
The constant-time algorithms in \cite{Porikli2008,Chaudhury2011,Chaudhury2011a,Chaudhury2013} are particularly relevant to the present work. The authors here proceed by approximating the Gaussian range kernel using polynomial and trigonometric functions, and demonstrate how the bilateral filter can be decomposed into a series of spatial Gaussian filters as result. Now, since a Gaussian filter can be implemented in constant-time using separability and recursion \cite{Deriche1993}, the overall approximation can therefore be computed in constant-time. 

\subsection{Present Contribution}

In this paper, we propose a fast $O(1)$ algorithm for computing \eqref{BF} which is motivated by the line of work in \cite{Porikli2008,Chaudhury2011,Chaudhury2011a,Chaudhury2013}.
In particular, we present a novel approximation for the range term in \eqref{BF} that allows us to decompose the bilateral filter into a series of fast Gaussian convolutions.
The fundamental difference between the above papers and the present approach is that instead of approximating the Gaussian and then translating the approximation, we directly approximate the translated Gaussians using the so-called Gauss-polynomials. The advantages of the proposed approximation are the following: 

$\bullet$ It is generally much more accurate than the polynomial approximation in \cite{Porikli2008}.

$\bullet$  For a fixed approximation degree (to be defined shortly), it leads to exactly half the number of Gaussian filterings than that required by the approximation in \cite{Chaudhury2011,Chaudhury2013}, and hence has a smaller run time.

$\bullet$ It does not involve transcendental functions such as $\cos( \omega x)$ and $\sin( \omega x)$ which are used in  \cite{Chaudhury2011,Chaudhury2013}. It only involves polynomials (and just a single Gaussian) and hence can be efficiently implemented on hardware \cite{Bailey2011,Muller2006}. This is partly what motivated the present work.

Moreover, we also show how the proposed approximation can be improved by first centering the range data, then applying the approximation algorithm, and finally adding back the centre to the processed range data.

\section{Gauss-Polynomial Decomposition}
\label{sec:decomp}

The main idea in \cite{Porikli2008,Chaudhury2011} was to approximate the range kernel in \eqref{kernel} using appropriate polynomials and trigonometric functions.
By using these approximations in place of the Gaussian kernel, it was shown that the numerator and denominator of \eqref{BF} can be approximated using a series of Gaussian filtering.

\begin{figure}[!htb]
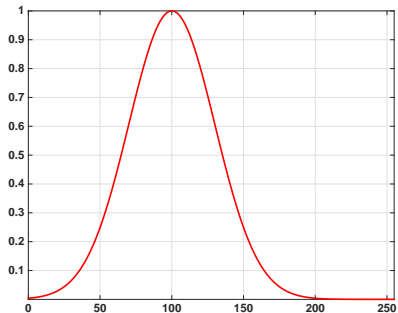
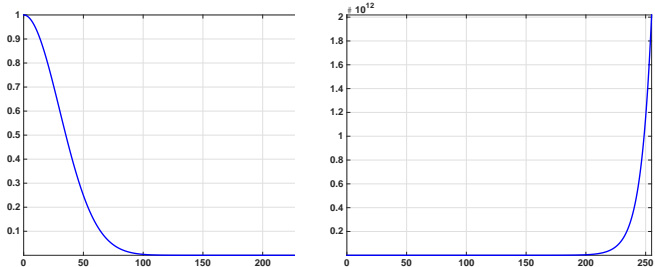

\centering
\subfloat[Gaussian centered at $\tau=100$.]{\includegraphics[width=0.6\linewidth]{./figures/originalGauss.eps}}  \\
\subfloat[Gaussian centered at $\tau=0$.]{\includegraphics[width=0.5\linewidth]{./figures/bell.eps}}  
\subfloat[Exponential function.]{\includegraphics[width=0.5\linewidth]{./figures/slope.eps}} 
\caption{An instance of the decomposition in \eqref{decomp} corresponding to $\tau=100$ and $\sigma_r=30$. We used $L=0$ and $U=255$ corresponding to the intensity range of an $8$-bit grayscale image. Up to a scaling factor of about 4e-3,  (a) is a product of  (b) and  (c).} 
\label{decomposition}
\end{figure}

The present idea is to consider the translates of the range kernel $g_{\sigma_r}(t - \tau)$ that appear in \eqref{BF}, where $t=f(\i-\j)$ and $\tau=f(\i)$ take values in some intensity range, say, $[L,U]$. For example, $L=0$ and $U=255$ for an $8$-bit grayscale image. We can write 
\begin{equation}
\label{decomp}
g_{\sigma_r}(t - \tau) = \exp\left(- \frac{\tau^2}{2\sigma_r^2}\right) \exp\left(- \frac{t^2}{2\sigma_r^2}\right) \exp\left(\frac{\tau t}{\sigma_r^2}\right).
\end{equation}
For a fixed translation $\tau$, this is a function of $t$. Notice that the first term is simply a scaling factor, while the second term is a Gaussian centered at the origin. In fact, the second term essentially contributes to the ``bell'' shape of the translated Gaussian. The third term is a monotonic exponential function which is increasing or decreasing depending on the sign of $\tau$. This term helps in translating the Gaussian to $t=\tau$. The decomposition in \eqref{decomp} plays an important role in the rest of the discussion and is illustrated in Figure \ref{decomposition}. 

Consider the Taylor series of the exponential function, 
\begin{equation}
\label{approx}
\exp\left(\frac{\tau t}{\sigma_r^2}\right) = \sum_{n=0}^N \frac{1}{n!}  \Big(\frac{\tau t}{\sigma_r^2}\Big)^n + \text{ higher-order terms}.
\end{equation}
By dropping the higher-order terms, we obtain the following approximation of \eqref{decomp}:
\begin{equation}
\label{GaussPolynomial}
 \exp\left(- \frac{\tau^2}{2\sigma_r^2}\right) \exp\left(- \frac{t^2}{2\sigma_r^2}\right) \Bigg[\sum_{n=0}^N \frac{1}{n!}  \Big(\frac{\tau t}{\sigma_r^2}\Big)^n \Bigg].
\end{equation}
Being the product of a Gaussian and a polynomial of degree $N$, we will henceforth refer to \eqref{GaussPolynomial} as a ``Gauss-polynomial'' of degree $N$. 

\begin{figure}[!htb]
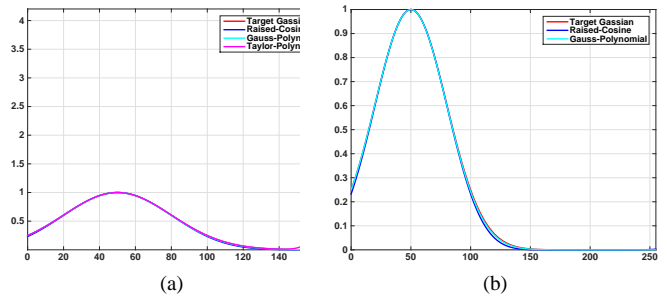

\centering
\subfloat[]{\includegraphics[width=0.5\linewidth]{./figures/compareRCandGPandTaylor.eps}}  
\subfloat[]{\includegraphics[width=0.5\linewidth]{./figures/compareRCandGP.eps}}  
\caption{Approximation of $g_{\sigma_r}(t - \tau)$ where $\sigma_r=30$ and $\tau=10$. The raised-cosine \cite{Chaudhury2011}, the Taylor polynomial \cite{Porikli2008}, and the Gauss-polynomial have the same degree $N=10$. In subfigure (a), notice how the Taylor polynomial quickly goes off to $+\infty$ as one moves away from the origin. For this reason, we restricted the plot to the interval $[0,170]$, although the desired approximation range is the full dynamic range $[0,255]$. The approximation over $[0,255]$ for the raised-cosine and the Gauss-polynomial are separately provided in subfigure (b).} 
\label{comparison}
\end{figure}

At this point, we note that one of the proposals in \cite{Porikli2008} was to approximate $g_{\sigma_r}(t - \tau)$ using its Taylor polynomial. 
The fundamental difference with our approach is that instead of approximating the entire Gaussian, we approximate one of its component, namely the monotonic exponential in \eqref{decomp}. The intuition behind this is that a polynomial eventually goes to infinity as one moves away from the origin. This makes it difficult to 
approximate the Gaussian function on its asymptotically-decaying tails. As against this, the exponential function in \eqref{decomp} is monotonic  and hence can be more accurately approximated using polynomials. This is explained with an example in Figure \ref{comparison}. In particular, notice in Figure \ref{comparison} (b) that the Gauss-polynomial approximation is fairly accurate over the entire range of interest and is comparable to the raised-cosine approximation \cite{Chaudhury2011}. 

\begin{figure}[!htb]
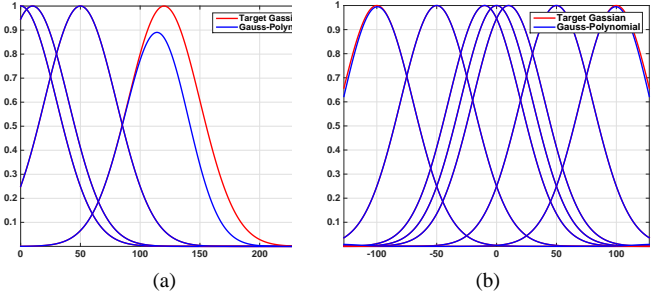

\centering
\subfloat[]{\includegraphics[width=0.5\linewidth]{./figures/differentTauOver0255.eps}}
\subfloat[]{\includegraphics[width=0.5\linewidth]{./figures/differentTaus.eps}}
\caption{(a) Approximation of $g_{\sigma_r}(t - \tau), \sigma_r=30,$ over $[0,255]$ using Gauss-polynomials of degree $20$. We vary $\tau$ over $0,10,50,120$. (b) Approximation over the centered interval $[-255/2,255/2]$, where we vary $\tau$ over $-100,-50,0,10,50,100$. This demonstrates how the approximation accuracy can be improved by simply centering the range interval at the origin.} 
\label{comparisonTau}
\end{figure}

We note that for a fixed degree $N$, the accuracy of the Gauss-polynomial approximation in \eqref{GaussPolynomial} depends on $\tau$. Indeed, when $\tau=0$, there is nothing to approximate since the exponential function reduces to a constant in this case. On the other hand, we see from \eqref{GaussPolynomial} that the magnitude of the higher-order terms increases with increase in $|\tau|$, and the approximation accuracy drops as a result. This is demonstrated with an example in Figure \ref{comparisonTau} (a). Of course, the approximation can be improved by using Gauss-polynomials of higher degree. However, we note that for a fixed degree,  the approximation accuracy can be improved simply by reducing the maximum $|\tau|$ that appears in \eqref{GaussPolynomial}. We propose the following ``centering'' trick in which we set, for example,  
\begin{equation*}
t_c = \textrm{mean} \ \{ f(\i) : \i \in I\} = \frac{1}{|I|} \sum_{\i \in I} f(\i).
\end{equation*}
We next translate each pixel intensity by $t_c$, which has the effect of centering the transformed intensity range at the origin. For example, when $L=0$ and $U=255$, the maximum $|\tau|$ equals $255$. However, if we center the intensity range  $[0,255]$, say at $t_c=100$, then we can effectively reduce the maximum $|\tau|$ to $155$. The Gauss-polynomial approximations obtained after the centering are shown in Figure \ref{comparisonTau} (b). The  above idea of centering is compatible with the bilateral filter precisely because of the following property of the bilateral filter.

\begin{proposition}
If $h(\i) = f(\i) - t_c$, then 
\begin{equation}
\label{centering}
f_{\mathrm{BF}}(\i) = h_{\mathrm{BF}}(\i) + t_c \qquad (\i \in I). 
\end{equation}
\end{proposition}
This is a simple consequence of the fact that the range kernel depends only on the intensity difference, and that for a fixed range term, \eqref{BF} preserves constant functions. In other words, we can first centre the intensity range, apply the bilateral filter, and finally add back the centre to the output.

\section{Fast Bilateral Filter}
\label{sec:fast}

We now present the constant-time implementation of \eqref{BF} using Gauss-polynomials. Suppose that $N$ is the degree of the polynomial in \eqref{GaussPolynomial}. For $n=0,\ldots,N+1$, define the images 
\begin{equation}
\label{intImg}
G_n(\i) =  \left(\frac{f(\i)}{\sigma_r}\right)^n \ \ \text{and } \ \ F_n(\i) = \exp\left(- \frac{f(\i)^2}{2\sigma_r^2}\right) G_n(\i).
\end{equation}
Denote the Gaussian filtering of $F_n(\i)$ by $\bar{F}_n(\i)$, that is,
\begin{equation}
\label{GaussFilter}
\bar{F}_n(\i) = \left(F_n \ast  g_{\sigma_s}\right)(\i) = \sum_{\j \in \Omega} g_{\sigma_s}(\j) F_n(\i - \j).
\end{equation}
Substituting $t=f(\i-\j)$  and $\tau=f(\i)$, and using the Gauss-polynomial approximation \eqref{GaussPolynomial} in place of $g_{\sigma_r}(t - \tau)$, it can be verified that (after interchanging  summations) we can express the numerator of \eqref{BF} as 
\begin{equation*}
\exp\left(- \frac{f(\i)^2}{2\sigma_r^2}\right)P(\i),
\end{equation*}
where
\begin{equation}
\label{P}
P(\i) = \sigma_r  \sum_{n=0}^N \frac{1}{n!} G_n(\i) \bar{F}_{n+1}(\i).
\end{equation}
Similarly, we can express the denominator of \eqref{BF} as
\begin{equation*}
\exp\left(- \frac{f(\i)^2}{2\sigma_r^2}\right)Q(\i),
\end{equation*}
where
\begin{equation}
\label{Q}
Q(\i) =  \sum_{n=0}^N \frac{1}{n!} G_n(\i) \bar{F}_{n}(\i).
\end{equation}
In other words, we can approximate \eqref{BF} by 
\begin{equation}
\label{approxBF}
\hat{f}_{\mathrm{BF}}(\i)=P(\i)/Q(\i).
\end{equation}
Note that we have effectively transferred the non-linearity of the bilateral filter to the intermediate images in \eqref{intImg}, which are obtained from the input image using  pointwise non-linear transforms. The main leverage that we get from the above manipulation is that, for any arbitrary $\sigma_s$, \eqref{GaussFilter} can be computed using $O(1)$ operations per pixel  \cite{Deriche1993}. The overall cost of computing \eqref{approxBF} is therefore $O(1)$ per pixel. In other words, we have a constant-time approximation of the bilateral filter. In this regard, we note that the above analysis holds if we replace the spatial Gaussian filter by any other filter (e.g., a box filter) that has a constant-time implementation.

\begin{algorithm}
\KwData{Image $\{f(\i) : \i \in I\}$, and parameters $\sigma_s,\sigma_r,N$.}
\KwResult{Approximation $\{\hat{f}_{\mathrm{BF}}(\i) : \i \in I\}$.}
$t_c = \textrm{mean} \ \{ f(\i) : \i \in I\}$\;
$h(\i) = f(\i) - t_c$\;
$G(\i) =1$\; 
$F(\i) = \exp(-h(\i)^2 / 2\sigma_r^2)$\;
$P(\i)=0$\;
$Q(\i)=0$\;
$H(\i)=h(\i)/\sigma_r$\;
$\bar{F}(\i) = \left(F \ast g_{\sigma_s} \right) (\i)$\;
$c=1$\;
\For{$n=0,1,\ldots,N$}{
$Q(\i)= Q(\i) + c \cdot G(\i) \bar{F}(\i)$\;
$F(\i) = H(\i)  F(\i)$\;
$\bar{F}(\i) = \left(F \ast g_{\sigma_s} \right) (\i)$\;
$P(\i) = P(\i) + c \cdot G(\i) \bar{F}(\i)$\;
$G(\i) = H(\i) G(\i)$\;
$c = c/(n+1)$\;
}
$\hat{f}_{\mathrm{BF}}(\i)= \sigma_r  \left(P(\i)/Q(\i)\right) + t_c$\;
\caption{Gauss-Polynomial Bilateral Filter (\texttt{GPF}).}
\label{GPF}
\end{algorithm}

The  overall algorithm is summarized in Algorithm \ref{GPF}. We will henceforth refer to this as the Gauss-Polynomial-based Bilateral Filter (\texttt{GPF}). Notice that we use centering and \eqref{centering} to improve the accuracy. Moreover, we efficiently implement steps \eqref{intImg} to \eqref{Q}. In particular, we recursively compute the images in \eqref{intImg} and the factorials in \eqref{P}  and \eqref{Q}. Notice that steps $2$-$7$, $11$-$12$, $14, 15$,  and $18$ are applied to each pixel (cheap pointwise operations). To avoid confusion, we note that the specification of the some of quantities in Algorithm \ref{GPF} are somewhat different from the corresponding definitions in \eqref{intImg} - \eqref{approxBF}.

It is clear that the main computations in \texttt{GPF} are the Gaussian filterings in step $13$ (and the initial filtering in step $8$). That is, the overall cost is dominated by the cost of computing $N+1$ Gaussian filterings. In this regard, we note that for the same degree $N$, the number of  Gaussian filterings required in \cite{Chaudhury2011} is $4(N+1)$. Indeed, we will see in Section \ref{sec:results} for a fixed $N$, the overall run-time of \texttt{GPF} is about a third of that of \cite{Chaudhury2011}. Finally, we note that  \texttt{GPF} involves the evaluation of a transcendental function just once in step 4. Thus, \texttt{GPF} is better suited to hardware implementation \cite{Bailey2011,Muller2006} compared to the algorithm in \cite{Chaudhury2011} which involves the repeated evaluation of cosine and sine functions.

\section{Experiments}
\label{sec:results}

\begin{figure}[!htb]
\centering
\subfloat[Test Image ($256 \times 256$).]{\includegraphics[width=0.48\linewidth]{./figures/peppers.eps}}  
\subfloat[Exact Implementation.]{\includegraphics[width=0.48\linewidth]{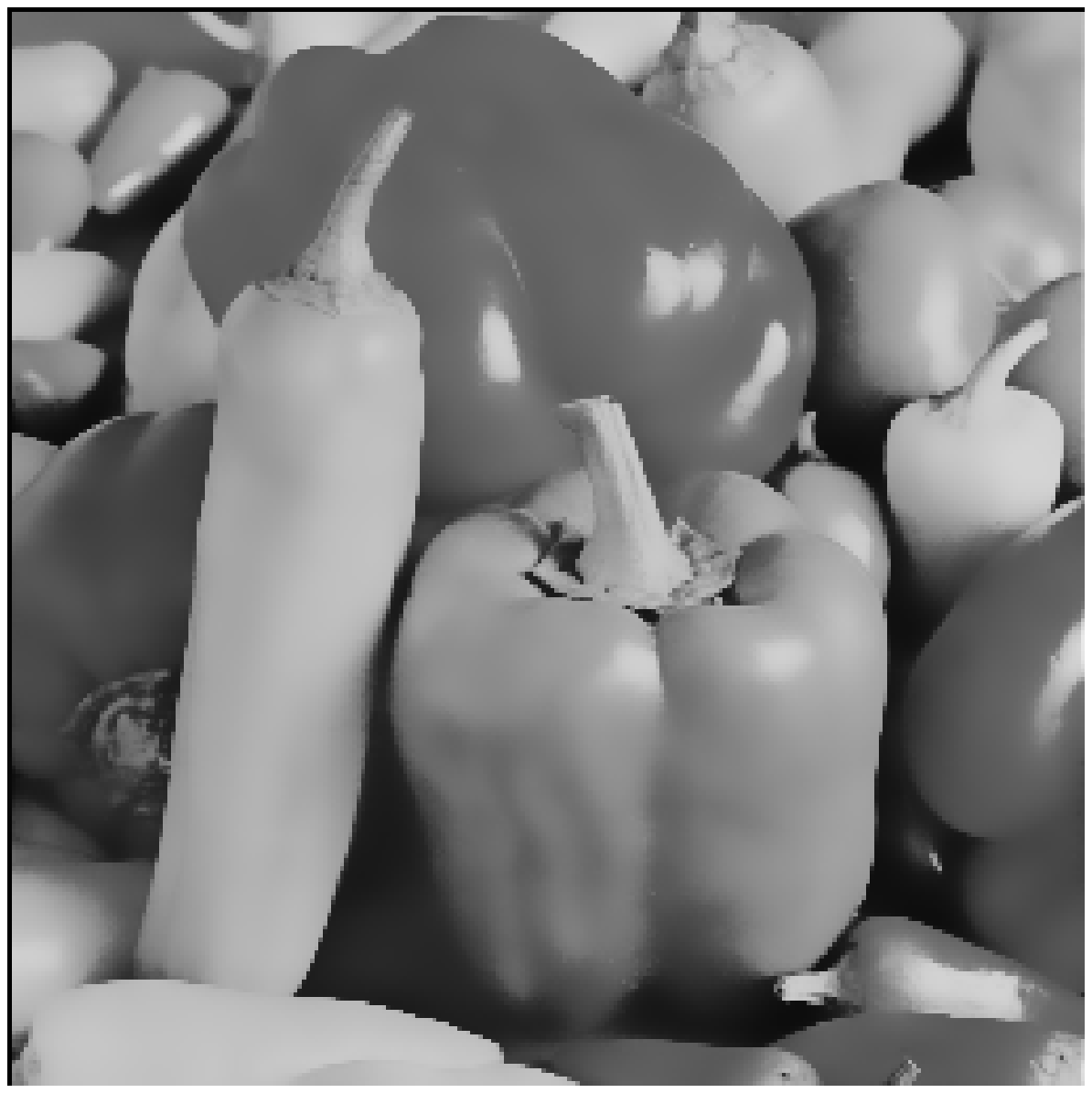}}  \\ \vspace{-1em}
\subfloat[Bilateral Grid \cite{Paris2006}, \textbf{7.8} dB.]{\includegraphics[width=0.48\linewidth]{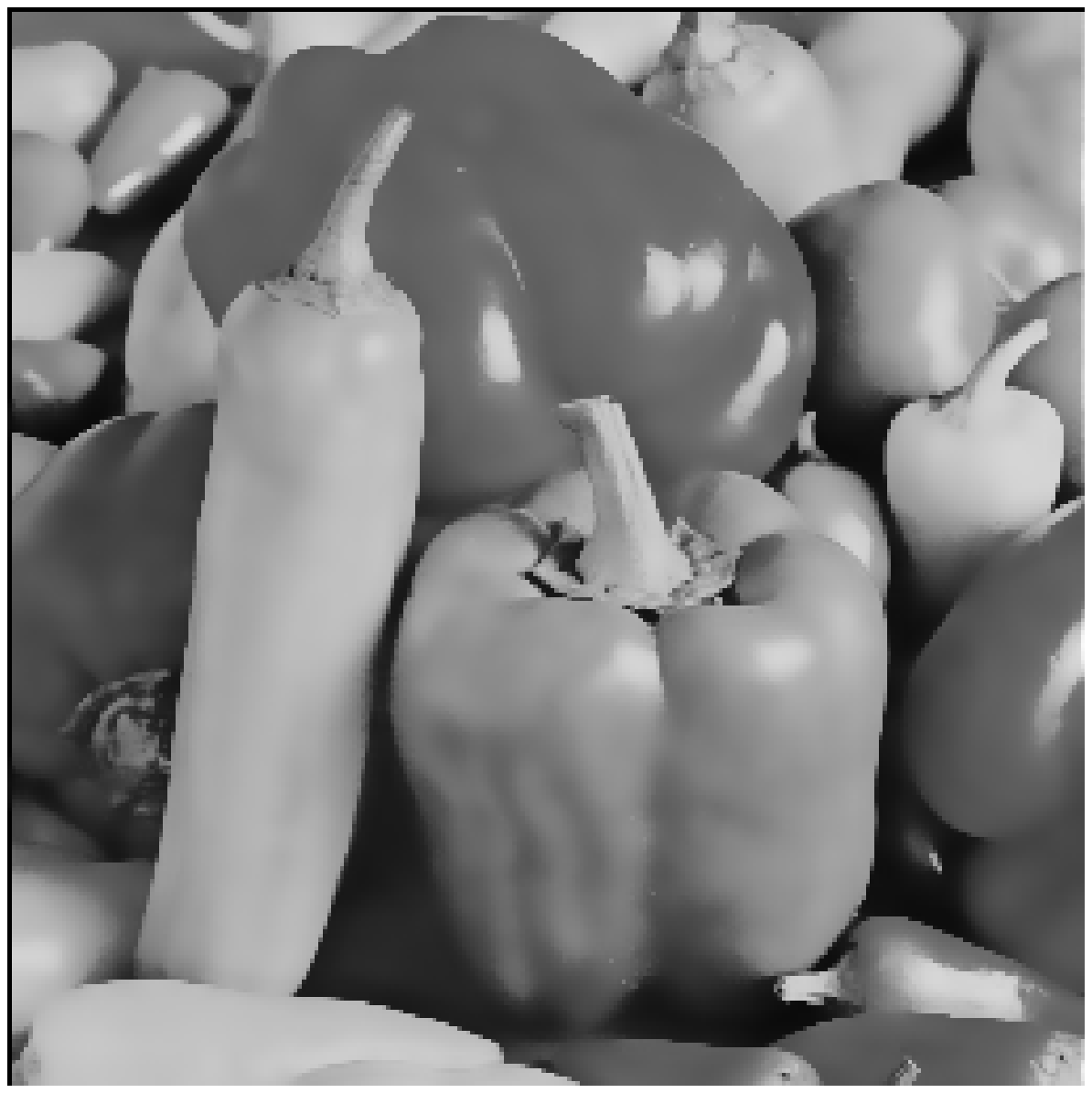}}  
\subfloat[Taylor-Polynomial \cite{Porikli2008}, \textbf{31} dB.]{\includegraphics[width=0.48\linewidth]{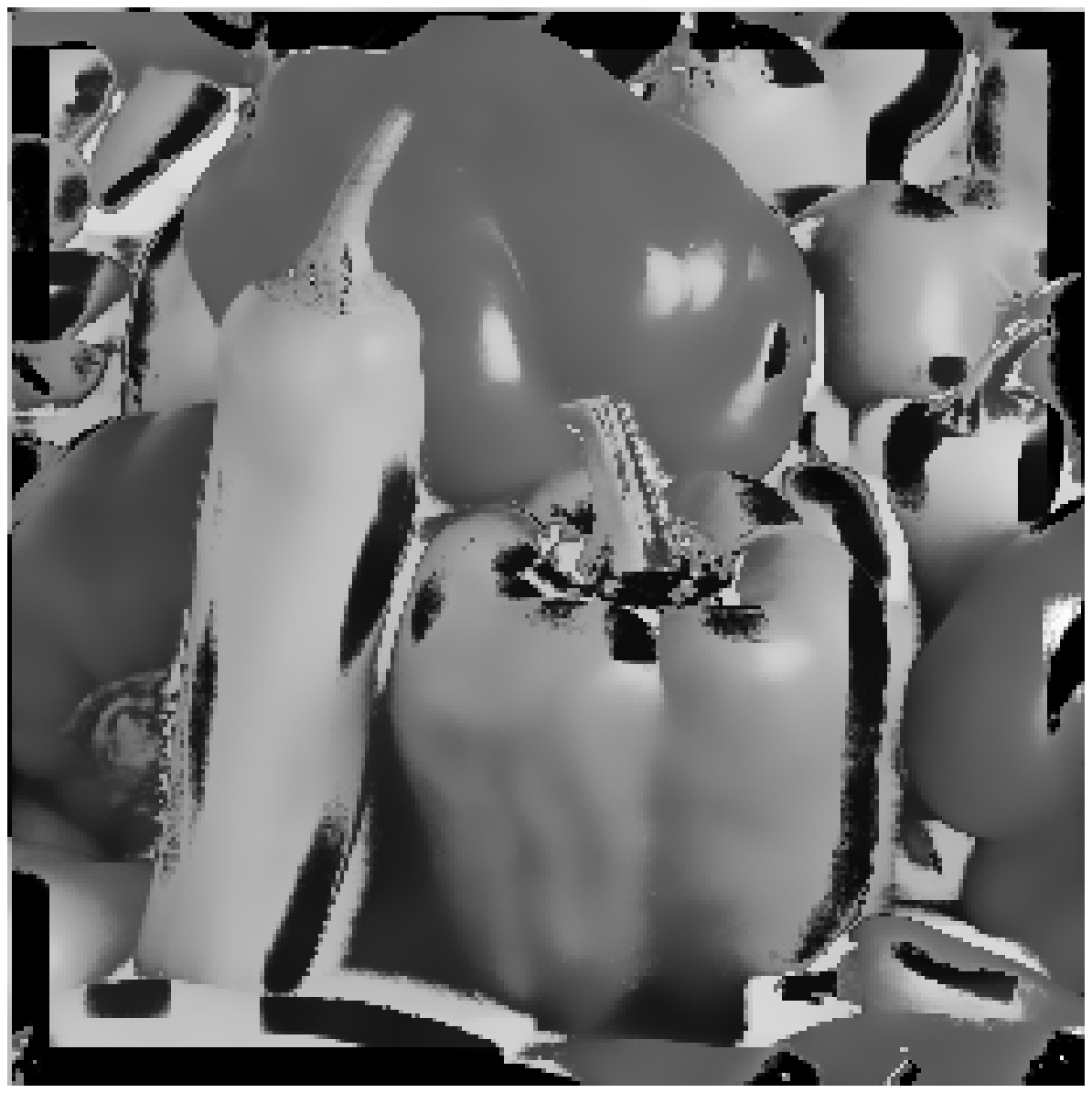}}  \\ \vspace{-1em}
\subfloat[Range Interpolation \cite{Yang2009}, \textbf{-1.1} dB.]{\includegraphics[width=0.48\linewidth]{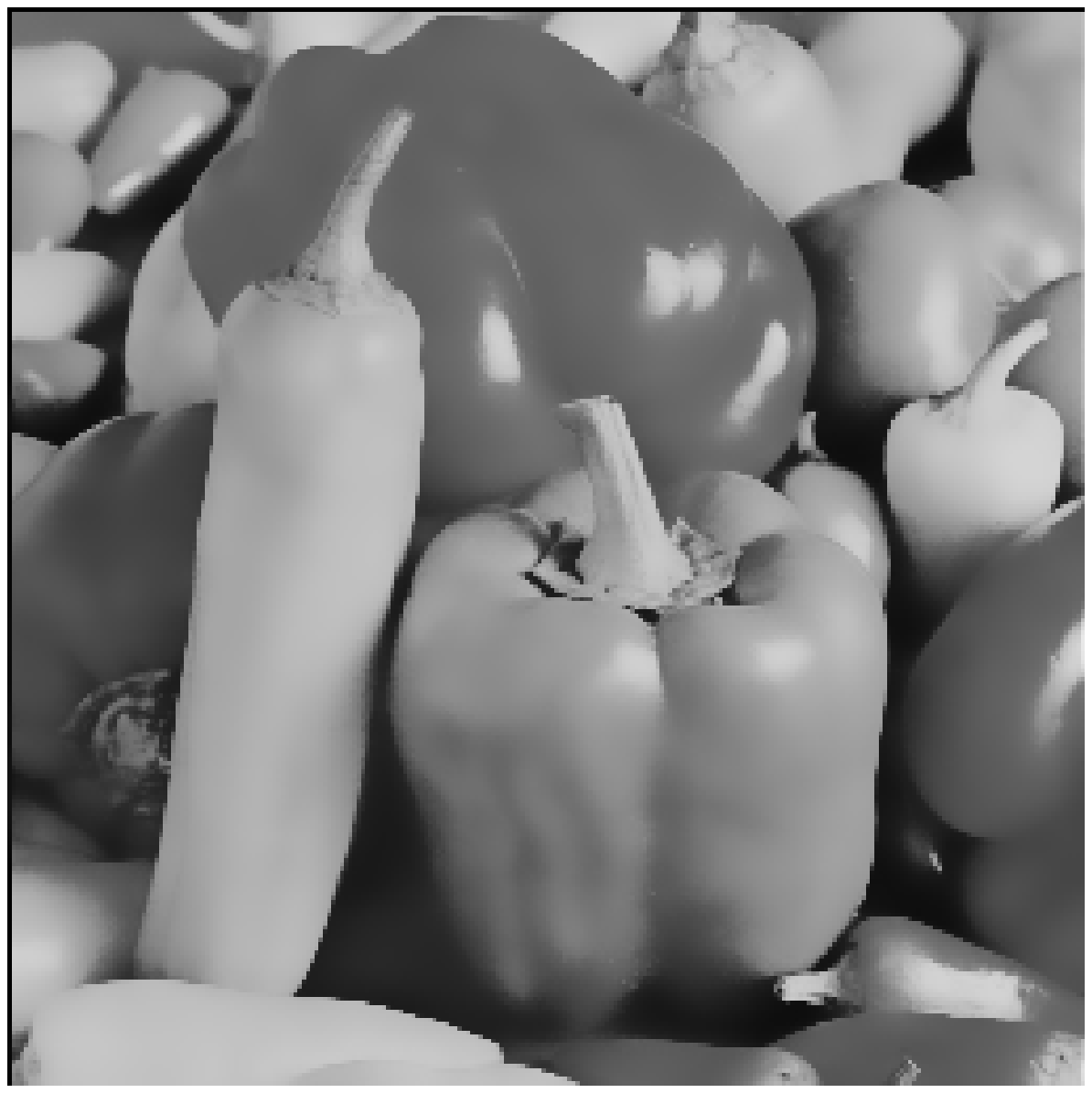}}  
\subfloat[Proposed (\texttt{GPF}), \textbf{-5.6} dB.]{\includegraphics[width=0.48\linewidth]{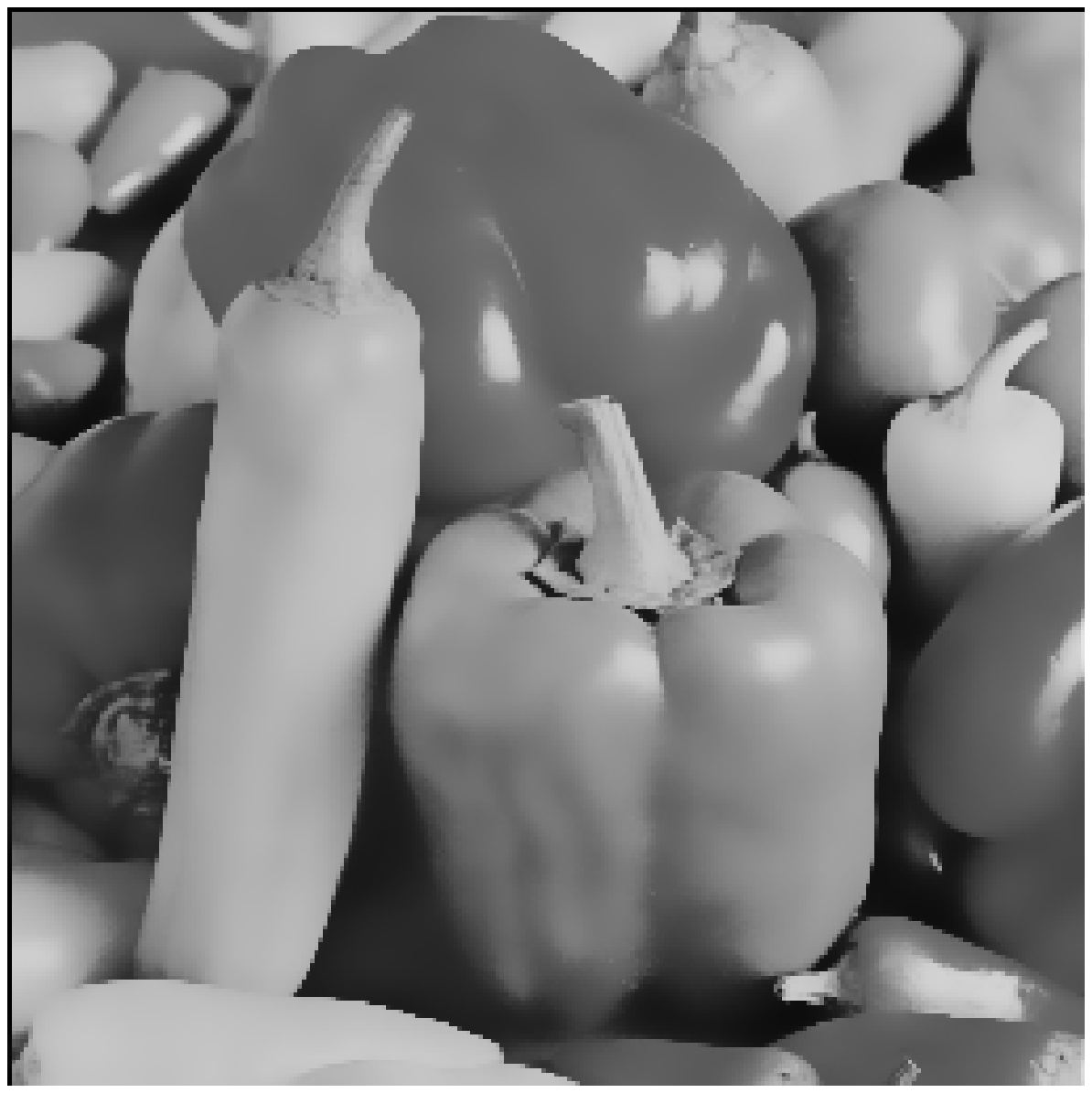}}  
\caption{Comparison of the exact implementation of \eqref{BF} with various fast algorithms for $\sigma_s=3,\sigma_r=30$. The degree of the respective functions (raised-cosine, Taylor polynomial, and Gauss-polynomial) used to approximate the range kernel is $20$. The MSE between the exact implementation and the various approximations are shown in bold.} 
\label{OutputImages}
\end{figure}

We now present some results concerning the accuracy and run-time of the proposed \texttt{GPF} algorithm. In particular, we compare it with some of the fast algorithms \cite{Paris2006,Porikli2008,Yang2009,Chaudhury2013}. The experiments were performed using Matlab on an Intel quad-core 2.7 GHz machine with 8 GB memory. We implemented the Gaussian filtering in \texttt{GPF} and \cite{Yang2009,Chaudhury2013} using Deriche's constant-time algorithm \cite{Deriche1993}. 
The average run-times of the various fast algorithms are reported in Table \ref{table} for a $256 \times 256$. We do not redundantly report the run-times for different image sizes, since this can roughly be estimated from the run-times in Table \ref{table} (the algorithms scale linearly with the number of pixels). 
Notice that the run time of \texttt{GPF} and \cite{Yang2009,Chaudhury2013} does not change appreciably with $\sigma_s$.
To evaluate the accuracy, we also report the mean-squared-error (MSE) between the exact implementation of \eqref{BF} and the result obtained using the fast algorithms in Table \ref{table}. In particular, the MSE between two images $f(\i)$ and $g(\i)$ is defined to be $10 \log_{10}(\text{MSE})$ dB, where $\text{MSE}=(1/|I|)\sum_{\i \in I}(f(\i)-g(\i))^2$.  
Notice that \texttt{GPF} is competitive with the existing algorithms in terms of accuracy and run-time. In particular, \texttt{GPF} has the smallest run-time, and its MSE is in general better than the rest of the algorithms and comparable to that of the raised-cosine-based approximation in \cite{Chaudhury2013}. 
The degree of the raised-cosine and the Gauss-polynomial filter is $20$ for all the experiments (this gives a good tradeoff between accuracy and run-time). 
In this regard, an open question is how the accuracy of \texttt{GPF} varies with the degree, as a function of $\sigma_s$ and $\sigma_r$. 
This will be addressed in future work.
Note that the run-time of the polynomial approximation in \cite{Porikli2008} is almost identical to that of the proposed algorithm and is hence not reported.
For a visual comparison, we report the result obtain on the \textit{Peppers} image in Figure \ref{OutputImages}.
Notice the visible distortions in subfigure (d) which arises on account of the poor approximation of the Gaussian kernel using Taylor polynomials.

\begin{table}[!htb]
\caption{The top table comparison of the average run-time of the exact, the proposed, and the fast algorithms in \cite{Paris2006,Yang2009,Chaudhury2013} for different values of $\sigma_s$ and fixed $\sigma_r=30$. We used the \textit{Peppers} image shown in Figure \ref{OutputImages}. 
The settings suggested  in \cite{Paris2006,Yang2009,Chaudhury2013} were used for the corresponding implementations.
The algorithms were implemented using Matlab on an Intel quad-core 2.7 GHz machine with 8 GB memory. 
The bottom table compares the MSE between the exact implementation of \eqref{BF} and the respective algorithms.}  
\vspace{2mm}
\centering 
\begin{tabular}{l*{6}{c}r}
\hline
\multicolumn{6}{c}{\textbf{Run-Time}} \\
\hline
     &       $2$  & $3$ & $4$  & $5$ & $10$  & $15$   \\
\hline
Exact          &1.5s    & 3.2s   & 5.3s     & 8.4s    &32.5s   & 73.2s    \\
\cite{Paris2006}       & 93ms     & 134ms    & 191ms  & 261ms      & 847ms       & 1.92s   \\
\cite{Yang2009}           & 112ms         & 118ms    & 115ms  & 116ms      & 118ms       & 120ms    \\
\cite{Chaudhury2013}           & 210ms         & 215ms    & 220ms  & 225ms      & 230ms       & 250ms    \\
\texttt{GPF}      & 74 ms         & 82ms    & 88ms  & 89ms      & 95ms       & 98ms    \\

\hline
\hline
\multicolumn{6}{c}{\textbf{MSE (dB)}} \\

\hline

\cite{Paris2006}       & 5.9     & 7.8    & 9.1  & 9.8      & 12.2       & 13.1   \\
\cite{Yang2009}           & -3.3         & -1.1    & 0.5 & 1.8      & 6.2       & 9.2    \\
\cite{Chaudhury2013}           & -10.5         & -6.4    & -3.8  &  -1.7      & 4.4       & 7.8    \\
\texttt{GPF}      & -9.6         & -5.6    & -3.1  & -1.1      & 5.1       & 8.4    \\

\hline
\end{tabular}
\label{table}
\end{table}

\section{Conclusion}
\label{sec:conc}

We presented a fast algorithm for bilateral filtering based on Gauss-polynomial decompositions of the translations of the range kernel. We presented some preliminary results which demonstrated the accuracy and speed of the algorithm in comparison to some of the existing fast algorithms for bilateral filtering. In particular, we saw that the algorithm is comparable to the one in \cite{Chaudhury2013}, but has a smaller run time (about a third). Moreover, as remarked in the introduction, the proposed algorithm has an edge over  \cite{Chaudhury2013} in the context of hardware implementation -- it is based on polynomials and does not involve the computation of multiple trigonometric functions \cite{Muller2006}. We note that the algorithm has a direct extension to other variants of the bilateral filter including the joint and guided filter \cite{GuidedBF,JointBF}, and can also be extended for handling volume and video data \cite{Adams2010}. 

\vfill\pagebreak

\bibliographystyle{IEEEbib}

\end{document}